\documentclass[letterpaper, 10 pt, conference]{ieeeconf}  

\IEEEoverridecommandlockouts                              

\usepackage{url}

\usepackage{makeidx}         
\usepackage{graphicx}        
\usepackage{multicol}        
\usepackage{amsmath}
\usepackage{amssymb}
\usepackage{booktabs}
\usepackage{xspace}
\usepackage{color}
\usepackage[hang,flushmargin]{footmisc}
\usepackage{microtype}
\usepackage{float}
\usepackage{array}
\usepackage{multirow}
\usepackage{flushend}
\usepackage{cite}

\usepackage{booktabs}
\usepackage{xspace}
\usepackage{array}
\usepackage{tabularx}
\let\labelindent\relax
\usepackage{enumitem}
\usepackage{siunitx}
\usepackage{gensymb}
\usepackage{microtype}
\usepackage{makecell}
\usepackage{threeparttable}

\usepackage{csquotes}
\usepackage[breaklinks,colorlinks]{hyperref}

\newcommand{\figref}[1]{Fig.~\ref{#1}}
\newcommand{\tabref}[1]{Tab.~\ref{#1}}

\usepackage{cite}
\usepackage{amsmath,amssymb,amsfonts}
\usepackage{algorithmic}
\usepackage{graphicx}
\usepackage{textcomp}
\usepackage{xcolor}

\usepackage{booktabs}
\usepackage{outlines}
\usepackage[hang,flushmargin]{footmisc}
\usepackage{stmaryrd}

\usepackage{microtype}
\usepackage{hhline}
\usepackage{multirow}
\usepackage{siunitx}
\usepackage{makecell}
\usepackage{gensymb}
\usepackage{placeins}

\definecolor{cvprblue}{rgb}{0.21,0.49,0.74}
\usepackage[breaklinks,colorlinks]{hyperref}

\usepackage[capitalize]{cleveref}
\crefname{section}{Sec.}{Secs.}
\Crefname{section}{Section}{Sections}
\Crefname{table}{Table}{Tables}
\crefname{table}{Tab.}{Tabs.}
\crefname{algorithm}{Algo.}{Algos.}
\Crefname{algorithm}{Algorithm}{Algorithms}

\definecolor{red}{RGB}{255, 0, 0}   
\definecolor{orange}{RGB}{255, 77, 0}   
\definecolor{green}{RGB}{0, 128, 0}   
\definecolor{purple}{RGB}{160, 32, 240}   
\definecolor{lightblue}{RGB}{52, 155, 235}   
\definecolor{darkmagenta}{RGB}{204, 51, 139} 

\def\BibTeX{{\rm B\kern-.05em{\sc i\kern-.025em b}\kern-.08em
    T\kern-.1667em\lower.7ex\hbox{E}\kern-.125emX}}

\usepackage{pifont}

\def\ourmodel{AmbResVLM\xspace}

\title{\LARGE \bf
    Robotic Task Ambiguity Resolution via Natural Language Interaction
}

\author{Eugenio Chisari$^{1}$, Jan Ole von Hartz$^{1}$, Fabien Despinoy$^{2}$, and Abhinav Valada$^{1}$
\thanks{$^{1}$ Department of Computer Science, University of Freiburg, Germany.}%
\thanks{$^{2}$ Toyota Motor Europe.}%
\thanks{This work was funded by Toyota Motor Europe.}
}
    
\begin{document}

\maketitle

\begin{abstract}
Language-conditioned robotic policies allow users to specify tasks using natural language. 
While much research has focused on improving the action prediction of language-conditioned policies, reasoning about task descriptions has been largely overlooked.
Ambiguous task descriptions often lead to downstream policy failures due to misinterpretation by the robotic agent.
To address this challenge, we introduce \ourmodel, a novel method that grounds language goals in the observed scene and explicitly reasons about task ambiguity.
We extensively evaluate its effectiveness in both simulated and real-world domains, demonstrating superior task ambiguity detection and resolution compared to recent state-of-the-art methods.
Finally, real robot experiments show that our model improves the performance of downstream robot policies, increasing the average success rate from 69.6\% to 97.1\%.
We make the data, code, and trained models publicly available at \url{https://ambres.cs.uni-freiburg.de}.

\end{abstract}

\section{Introduction}
\label{sec:intro}

Generalization to novel and diverse scenarios has long been a critical objective in robotic policy learning. 
A promising approach to achieve this goal is through natural language task specification, which offers a powerful and intuitive medium for human-robot interaction~\cite{honerkamp2024language,werby2024hierarchical,0_black_2024}. 
By leveraging the expressiveness and flexibility of human language, we can train generalist policies capable of handling a wide range of scenarios.

Language-conditioned policies must integrate information from multiple modalities, including the language task description, visual observations, and the appropriate actions required to accomplish the task. 
Given the success of Vision-Language Models (VLMs) in other domains, they have gained significant interest in robotics.
Using VLMs zero-shot~\cite{huang2023voxposer, duan2024manipulate} as well as fine-tuning them into Vision-Language-Action models (VLAs)~\cite{octo_2023, kim24openvla, 0_black_2024, wen2024tinyvla} has shown promising results in challenging tasks such as folding a t-shirt and wiping a table. 
These methods take advantage of the richness of web-scale data encoded in pre-trained foundation models to enhance the generalization capabilities of their proposed robot policies.
However, all aforementioned language-conditioned imitation learning policies are inherently reactive, mapping observations directly to actions without intermediate reasoning. 
To address this limitation, recent work has explored integrating planning~\cite{singh2023progprompt, liang2023code} and chain-of-thought reasoning~\cite{zawalski24ecot} into action generation, enabling more deliberate and thoughtful decision-making.

Despite these advancements, existing methods often assume that language goals are clear and well-defined. In unstructured real-world environments, however, language goals can be ambiguous, leading to policy failures. Ambiguous task descriptions represent a significant yet overlooked challenge, as they can result in misinterpretations by the agent and reduce the rate of successful task completion. 
To mitigate task failures, prior works have investigated how humans can provide corrective feedback to guide policy execution~\cite{chisari2021correct, liu2023robot, celemin2022interactive, shi2024yell} and how failure detection mechanisms can trigger retrial behaviors when the task outcome does not match the specified goal~\cite{du2023vision, duan2024aha, xiong2024aic}. 
While effective, these approaches employ reactive mechanisms that are only triggered after a failure has occurred or is about to occur.

\begin{figure}
    \centering
    \includegraphics[width=0.48\textwidth]{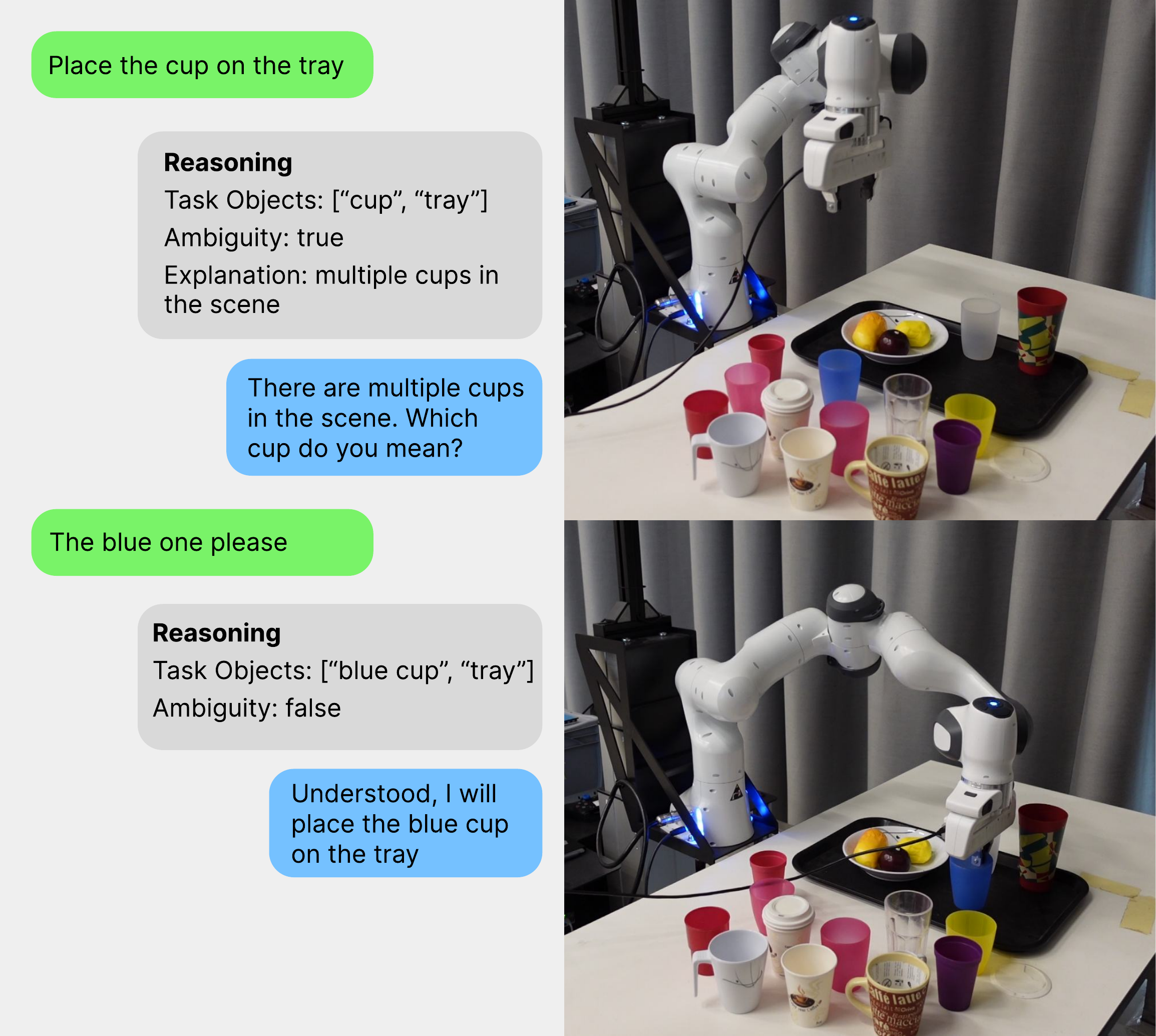}
    \caption{{\ourmodel reasons about the provided task description and grounds it in the observed scene. It then predicts whether the task is ambiguous and generates the appropriate query to the human user to disambiguate it. Finally, it interprets the user clarification, grounding the information again in the observed scene and enabling successful task execution.}}
    \label{fig:example}
\end{figure}

In this work, we introduce \ourmodel, a novel approach to address task ambiguity by grounding language goals in visual observations. Our model interprets the natural language task description within the context of the observed scene and assesses the clarity or ambiguity of the specified task. 
By reasoning about the task description itself, our approach preemptively identifies and resolves ambiguities, thereby preventing many failures that could arise from unclear goals.
When task ambiguity is detected, our agent actively queries a human user for clarification using natural language grounded in the observed scene. 
An illustration of the interactions that \ourmodel enables is visualized in Fig.~\ref{fig:example}.
Our model is able to formulate relevant questions, interpret human responses, and resolve ambiguities in the original task description. 
We evaluate \ourmodel on the ambiguity resolution task in two different domains, a simplified simulated domain and a real-world office environment. 
Furthermore, we conduct real-world robot experiment running ambiguity resolution alongside downstream manipulation policies. 
With this evaluation, we show that our model improves the robustness and reliability of downstream language-conditioned policies, increasing average success rate by 27.5 percentage points.

In summary, our main contributions are the following:
\begin{outline}
    \1 An automated pipeline to generate ambiguous image-task pairs, a dataset, and an evaluation protocol to test current and future models on task ambiguity resolution.
    \1 \ourmodel{}, a novel approach for resolving ambiguities in natural language robotic task descriptions.
    \1 Extensive experimental evaluations of the capabilities of our model, including evaluating its positive impact on downstream real-robot policy execution.
    \1 Code and pretrained models publicly available at \url{https://ambres.cs.uni-freiburg.de}.
\end{outline}
\section{Related Work}
\label{sec:related_work}

Prior works have extensively investigated how correction from expert users, failure detection, and ambiguity resolution can be adopted to mitigate common failure modes of robotic policies. In the following sections, we provide an overview of the most prominent methods.

{\parskip=2pt
\noindent\textit{Robot Actions Correction}: An effective method to improve the performance of robotic policies is by allowing expert users to directly provide corrective feedback to the robot~\cite{celemin2022interactive}.
Early works~\cite{chisari2021correct, liu2023robot} used common computer interfaces, such as keyboards and gamepads, to provide this information. 
More recently, natural language has emerged as a powerful and convenient modality that enables robots to understand both user feedback and reason about their actions.
Yell At Your Robot~\cite{shi2024yell} is a framework that enables agents to adapt to natural language feedback from users and incorporate this feedback into an iterative training process to improve future actions. 
For scenarios where a human supervisor is not available, RoboReflect~\cite{luo2025roboreflect} uses VLMs to enable self-reflection and autonomous error correction in object grasping tasks.}

With the widespread adoption of foundation models in robotic systems, task success/failure detection has increasingly gained importance.
A pioneering work in this field is SuccessVQA~\cite{du2023vision}, a VLM-based approach to detect task success in different domains, ranging from robotics to egocentric human videos. 
However, pure failure detection models are limited to a binary output and do not provide explanations about the task failure nor corrective actions to enable corrective behaviors. AHA~\cite{duan2024aha}, a vision language model, overcomes these limitations by detecting as well as reasoning over failures in robotic manipulation.
It is trained with instruction fine-tuning and can be used to improve task performance in all methods that use VLMs for reward generation and task planning.
Similarly, AIC~\cite{xiong2024aic} utilizes multi-modal large language models to correct SE(3) pose prediction failures of articulated objects.

While corrective feedback and failure detection are valuable for steering policies towards correct behaviors, they are reactive measures that come into play only after a mistake has occurred, or when it is about to occur. These methods also require constant human monitoring, which may not be always feasible.
On the other hand, the goal of \ourmodel is to detect and correct ambiguities in the language task description so that potential failures are prevented before they can happen.

{\parskip=2pt
\noindent\textit{Task Ambiguity}: Resolving ambiguous task descriptions is an important and often overlooked challenge when deploying intelligent robotic systems. 
An early work in this direction is presented in~\cite{dougan2022asking}, where the robot asks follow-up questions to resolve ambiguities in human-robot conversations. The method uses an off-the-shelf object detector to detect requested objects in the scene and then queries the user based on predicted confidence values.
More recently, KnowNo~\cite{knowno2023} leverages LLMs to reason about language task descriptions. The authors cast the ambiguity prediction problem as a question-answering problem. First, the LLM is prompted to predict a list of four robot actions to execute, labeled A) to D). Second, the LLM is queried to predict which of the proposed tasks to execute. By inspecting the logits of the tokens corresponding to the four letters A) to D), KnowNo estimates the validity of each proposed action. If more than one high-level action is assigned a high probability, a human query is generated.}

Building upon KnowNo, a few extensions have been proposed to tackle different aspects of task ambiguity in robotics. 
Both CLARA~\cite{park2023clara} and LAP~\cite{mullen2024lap} use the LLM to reason not only about task ambiguity but also about task feasibility.
Finally, Introspective Planning~\cite{liang2025introspective} proposes a systematic approach to reason about task compliance and safety, in addition to ambiguity. The method consists of constructing a knowledge base of human-selected safe behaviors, which are then retrieved during deployment.

While demonstrating interesting results, all aforementioned methods only use the image observation of the scene to query off-the-shelf object detectors, limiting the task ambiguity reasoning solely to the text modality. Therefore, these methods fail when encountering ambiguities based on object semantics not captured by the specific detector used.
On the other hand, in this work, we demonstrate that by leveraging state-of-the-art VLMs, we can approach the task ambiguity resolution problem from a multi-modal perspective, reasoning on both the language task description and the visual observation.

\section{Technical Approach}
\label{sec:technical_approach}

\begin{figure*}
    \centering
    \includegraphics[width=\textwidth]{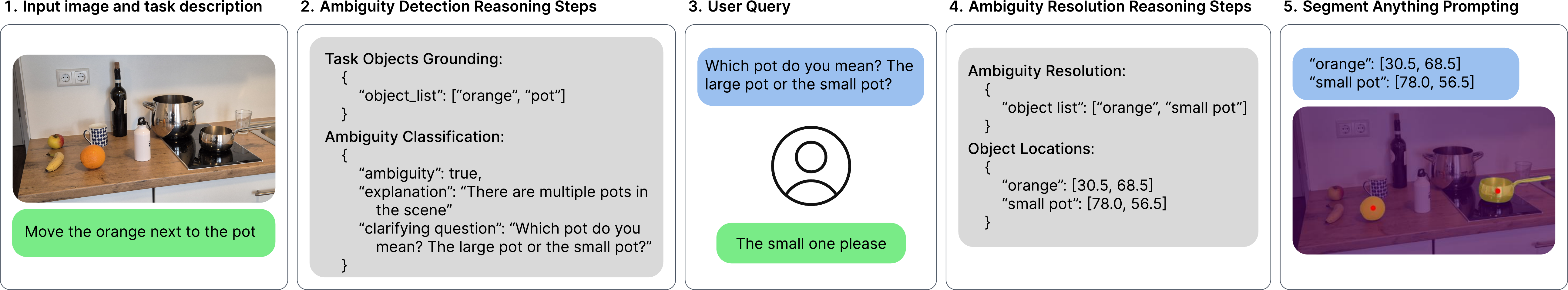}
    \caption{{Illustration of the reasoning process in \ourmodel. The model is provided with a task description in natural language and an image observation of the scene. The first reasoning step consists of grounding the task-relevant objects in the visual observation and classifying whether the task-image pair is ambiguous or not. If the task is deemed ambiguous, a user query is generated. The second reasoning step consists of interpreting the feedback from the user, grounding the now unambiguous task objects, and predicting their locations in the image. Optionally, the final step consists of using the predicted coordinates to prompt the SegmentAnything model~\cite{ravi2024sam2} to predict object masks.}}
    \label{fig:method}
\end{figure*}

Our objective is to resolve robotic task ambiguity by simultaneously reasoning about visual observations and language task descriptions. 
To achieve this goal, our model must overcome several challenges. 
First, it must be able to ground the task relevant objects from the language description into the scene. Second, it must detect potential ambiguity of the provided task description by reasoning about the presence or absence of multiple semantically similar objects in the image.
Finally, it must be able to generate the appropriate human query, as well as correctly interpret the human response, grounding it back into the observed scene.

Concretely, we fine-tune a pre-trained VLM on a small labeled dataset of ambiguous task descriptions and image pairs to perform a sequence of predefined reasoning steps.
While conceptually simple, the implementation of this approach requires careful investigation of multiple key points: 1) What pre-trained VLM is the most appropriate for this task? 2) What reasoning step do we implement to successfully be able to resolve ambiguities? 3) How do we ensure that the output structure of the VLM is consistent to allow downstream tasks to parse its information? 4) How do we fine-tune the model efficiently?
We discuss each of these points in the next sections.

\subsection{Choice of Pre-trained Base VLM}

Over the past few months, many VLMs have been developed. While powerful, we did not consider closed source models for our method, as using their APIs can prove expensive over time, as well as because they are less flexible to work with compared to open source models.
Currently, the most prominent open source VLMs are the following: Llava~\cite{liu2024improved} connects a pre-trained CLIP-ViT visual encoder with the Vicuna LLM using a simple projection matrix; Qwen2-VL~\cite{wang2024qwen2} employs a unified paradigm to process both images and videos; Pixtral~\cite{agrawal2024pixtral} adopts a vision encoder trained from scratch; Molmo~\cite{deitke2024molmo} is trained for unique visual skills such as pointing and counting, as well as charts and tables reading.
We build our proposed \ourmodel upon Molmo, given its unique object pointing skill, which will allow us to detect and ground objects in the image directly via the VLM, without requiring a separate open vocabulary object detector. 

\subsection{Reasoning Steps for Task Ambiguity Resolution}

Our objectives when defining the reasoning steps of our ambiguity resolution model are twofold. 
First, we need to enable the robot to ground its reasoning in both the observed image and the natural language descriptions provided by the user. At the same time, the model must be able to generate natural language queries intuitive for the user to understand and reply.
Second, the reasoning output needs to be structured such that downstream tasks, in our case, object mask prediction and policy execution, can reliably parse the predicted information. 

The full reasoning process of \ourmodel is illustrated in Fig.~\ref{fig:method}, and each step is detailed as following:
\begin{itemize}
  \item \textit{Task object grounding and ambiguity classification}. First, the model reasons about the provided task description and outputs a list of all task-relevant objects requested by the user.
  Second, the model needs to reason about the observed image to determine whether the task description is ambiguous or not. The final output is a boolean value. To encourage the model to provide justified classification outputs, we also require it to output an explanation for its decision.
  \item \textit{User query}. If the task description is deemed ambiguous, the model must now query the human to ask for clarification. This query must be in natural language to be easily understood by the user. In case the task description is considered unambiguous, a clarifying question is not required.
  \item \textit{Ambiguity resolution and object detection}. Once the human user provides a response to the query, the model must be able to interpret it and ground it in the observed scene. 
  The output of this step is again a list of task relevant objects, this time updated to include the additional information received.
  Once all task relevant objects are uniquely identified, the model needs to detect their precise coordinate locations in the image. 
  Instead of having to rely on a separate object detector, as most prior works do, we can directly output this information from the Molmo VLM. 
\end{itemize}

To allow downstream tasks to utilize this information effectively, we train the model to output all the described reasoning steps as JSON, a widely used data format that can be easily parsed.

\subsection{Structured Output}
\label{sec:structured}

While few-shot prompting and fine-tuning the VLM with JSON-formatted data provide an effective way to shape the output of the model, the predefined structure is not always guaranteed at inference time, potentially leading to downstream tasks failing to parse the predicted information.

To overcome this issue, we adopt the structured generation approach from~\cite{willard2023efficient}.
In this method, JSON schemas are translated into regular expressions that can be formulated as Finite State Machines (FSM). The FSM allows to construct an index of all non-zero probability tokens for each step of the LLM text generation process. The resulting algorithm does not add overhead in terms of compute time, as it has a time complexity of $O(1)$, and only a small overhead in terms of memory, which has complexity $O(N)$, where $N$ is the size of the vocabulary used, generally much smaller compared to the billions of parameters of the network.

In \ourmodel, we define two different JSON schemas for our model to follow, which can be used to guarantee our desired reasoning structure illustrated in Fig.~\ref{fig:method}. The object grounding schema consists of a list of strings, and can be used for both the task object grounding as well as for the ambiguity resolution steps. On the other hand, the ambiguity classification schema consists of a dictionary with three keys: ambiguity, explanation and clarifying question. Ambiguity is a boolean value, while explanation and clarifying questions are strings. 
This approach allows us to guarantee that our output will follow our pre-defined JSON schema for its structured output, which in turn guarantees that downstream tasks will be able to reliably parse the prediction of our model.

\subsection{Dataset and Model Fine-Tuning}
\label{sec:dataset}

\begin{figure}
\centering
\begin{tabular}{@{}c@{}c}
  \includegraphics[trim=0 0 0 36,clip,width=0.395\linewidth]{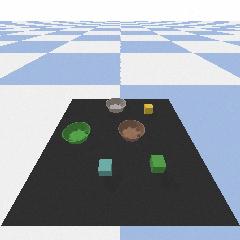}&
  \includegraphics[width=0.595\linewidth]{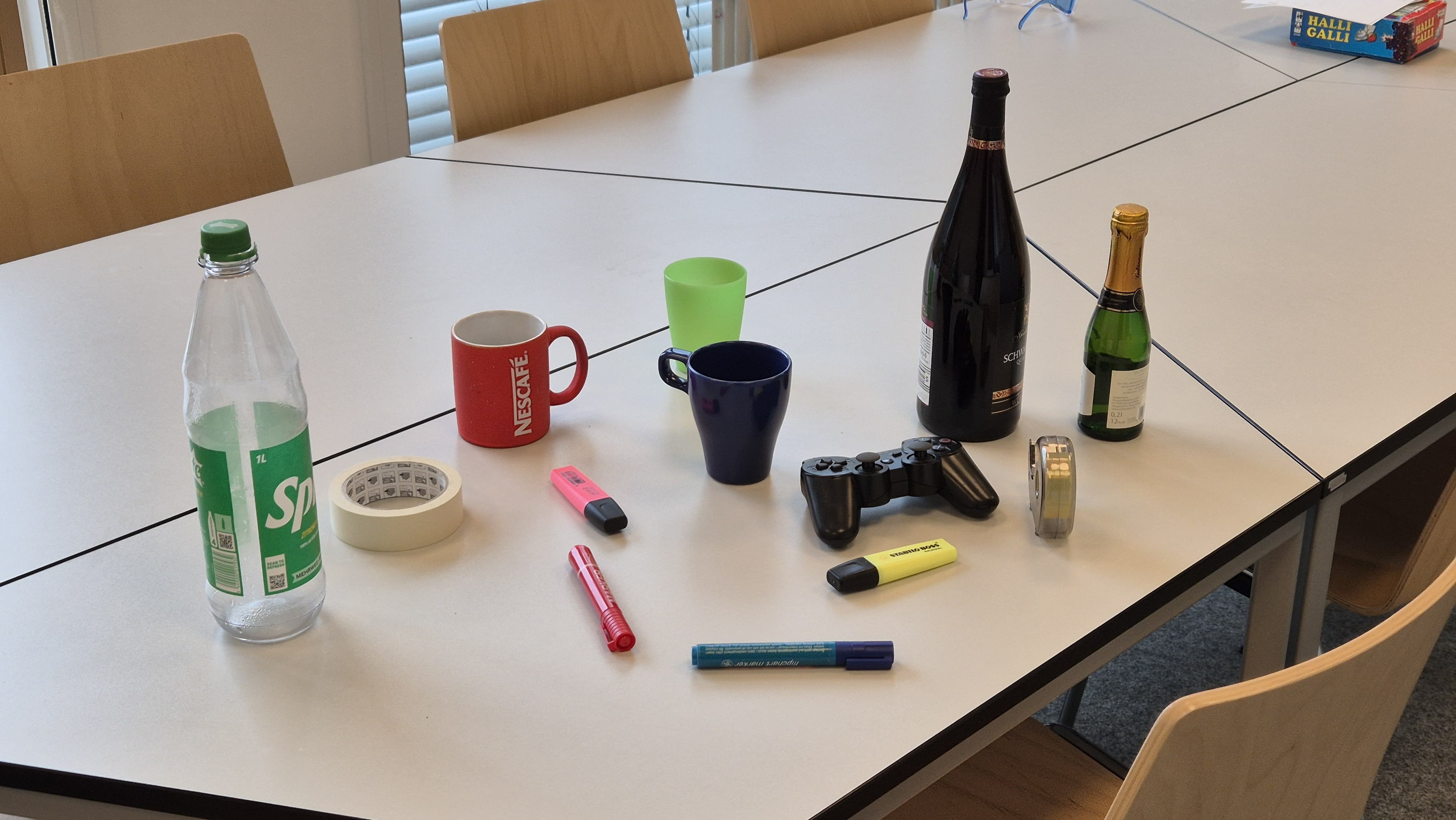}
\end{tabular}
\caption{Sample images from our dataset. Left: from the simulation environment. Right: from the real-world data.}
\label{fig:data}
\end{figure}

To both train and evaluate our model, we build two datasets of labeled image and task description pairs. 
The first dataset contains image-task pairs from a simplified simulated domain adapted from~\cite{knowno2023}. The scene consists of blocks and bowls of different colors placed on a table. 
The second dataset is built with real world images that we captured in an office and a kitchen environment, featuring a multitude of different objects. 
Samples from both datasets are depicted in Fig.~\ref{fig:data}.
Each dataset consists of 40 images. For each image, 20 task descriptions are generated, for a total of 800 image-task pairs. We sample both ambiguous and unambiguous task descriptions, so that the resulting dataset is balanced.
We perform a 50-50 train-test split to ensure enough samples for evaluation purposes.

We train our model from the pre-trained Molmo-7B-D-0924 VLM.
To enable efficient fine-tuning, we employ the Low-Rank Adaptation (LoRA)~\cite{hu2022lora} technique. LoRA freezes all pre-trained model weights and then injects trainable rank decomposition matrices, greatly reducing the total number of trainable parameters for fine-tuning. In particular, for \ourmodel we only fine-tune LoRA adapters for the language model parameters, while leaving the vision encoder untouched. Combined with training in bfloat16 format, this approach allows us to fine-tune the 7B Molmo VLM on a single RTX A6000 (48GB) GPU. We train our model using the AdamW~\cite{loshchilov2017decoupled} optimizer for one single epoch with the following hyperparameters: batch size of 2, learning rate of 0.0001, and weight decay of 0.01.

\section{Experimental Results}
\label{sec:experiments}

We design our evaluation to investigate the following research question: 1) Is \ourmodel able to ground task objects in the image? 2) How effectively can it detect ambiguity in the task description? 3) Is it able to reason about the user response and correctly resolve the original ambiguity? 4) Does downstream policy execution improve when tasks are disambiguated with \ourmodel?
We discuss each of these questions in the following subsections.

\subsection{Ambiguity Resolution}

\begin{table*}[t]
\centering
\caption{Performance comparison of the ambiguity resolution capabilities of \ourmodel with the KnowNo~\cite{knowno2023} baseline.}
\vspace{-0.2cm}
\label{tab:dataset_results}
\setlength{\tabcolsep}{3.2pt} 
\begin{threeparttable}
\begin{tabular}{l c c | c c c c c c c c |c c c c c c c c c}
\toprule
 &  && \multicolumn{7}{c}{ \textbf{Simulation}} && \multicolumn{7}{c}{ \textbf{Real World}} \\
\cmidrule{4-10} \cmidrule{12-18}
 &  && \multicolumn{1}{c}{Task Object} && \multicolumn{3}{c}{Ambiguity} && \multicolumn{1}{c}{Ambiguity} && \multicolumn{1}{c}{Task Object} && \multicolumn{3}{c}{Ambiguity} && \multicolumn{1}{c}{Ambiguity} \\
 &  && \multicolumn{1}{c}{Grounding} && \multicolumn{3}{c}{Classification} && \multicolumn{1}{c}{Resolution} && \multicolumn{1}{c}{Grounding} && \multicolumn{3}{c}{Classification} && \multicolumn{1}{c}{Resolution} \\
 \cmidrule{4-10} \cmidrule{12-18}
Model & Obs Type && IoU && Precision & Recall & F1 && Success && IoU && Precision & Recall & F1 && Success \\
\cmidrule{1-2} \cmidrule{4-4} \cmidrule{6-8} \cmidrule{10-10} \cmidrule{12-12} \cmidrule{14-16} \cmidrule{18-18}
KnowNo~\cite{knowno2023} & GT Obj List && --- && 0.49 & 0.96 & 0.64 && --- && --- && 0.47 & \textbf{0.81} & \textbf{0.60} && --- \\
AmbRes-prompt & RGB Image && \textbf{1.00} && 0.44 & 0.37 & 0.41 && \textbf{0.99} && 0.99 && 0.47 & 0.46 & 0.46 && \textbf{0.80} \\
AmbRes-finetune & RGB Image && \textbf{1.00} && \textbf{0.52} & \textbf{0.97} & \textbf{0.68} && \textbf{0.99} && \textbf{1.00} && \textbf{0.48} & 0.78 & 0.59 && 0.75 \\
\bottomrule
\end{tabular}
We evaluate with task-image pairs from both a simple simulated domain and a real world environment.
\end{threeparttable}
\end{table*}

Our goal is to evaluate the three main steps of our reasoning process: task object grounding, ambiguity classification, and ambiguity resolution. 
For this evaluation, we use the test split from the dataset we collected, as described in Sec.~\ref{sec:dataset}.
We compare \ourmodel with two baselines: KnowNo~\cite{knowno2023}, as well as an ablated version of our model which we call here \ourmodel-prompt. 
\ourmodel-prompt uses few-shot prompting instead of fine-tuning to steer the model output. The model is prompted with two examples from the dataset to provide it with context about the expected behavior.
Both models use the same structured output as described in Sec.~\ref{sec:structured}. A major difference of KnowNo compared to \ourmodel is that it requires the ground truth object list as input since it cannot reason over visual observations and operates with privileged information. 
Moreover, KnowNo does not reason about the human feedback. For this reason, we can only compare KnowNo to \ourmodel in the core capability of detecting task ambiguity. The results from this experiment are reported in Tab.~\ref{tab:dataset_results}. 

\textit{Task Object Grounding}. We first consider the task of object grounding. For this task, we evaluate the Intersection over Union (IoU) between the predicted set of objects and their ground truth.
The results show that both the fine-tuned and the few-shot prompting versions of \ourmodel achieve impressive results, grounding the requested task relevant objects correctly between 99\% and 100\% of the time. As previously stated, KnowNo does not provide this feature and therefore is excluded from this evaluation.

\textit{Ambiguity Classification}. We consider the ability of the model to detect ambiguities in the task description. To assess the classification performance, we use three common metrics: precision, recall, and F1. A notable result from this evaluation is that all three considered models have lower precision than recall, i.e., all models predict more false positives than false negatives. 
In other words, all models have a bias towards querying the user when in doubt. On one hand, this is positive, as it minimizes downstream task failures, but at the same time, it requires more effort than necessary from the user.
In our baseline comparison, we first observe that the fine-tuned model always outperforms the zero-shot prompted model. 
Second, we observe that \ourmodel achieves similar results to KnowNo, slightly higher in the simulation environment (0.68 vs 0.64 F1 score), and slightly lower in the real environment (0.59 vs 0.60 F1 score). 
The reason for this discrepancy is that the real-world domain is visually more complex. 
While our model processes image inputs, KnowNo operates with ground-truth object information, and therefore it is not affected by the increased visual complexity.
This result shows that \ourmodel achieves performance on par with the baseline that does not support visual observations and operates with privileged information instead.

\textit{Ambiguity Resolution}. We evaluate the ability of our models to interpret the feedback from the user and resolve the task ambiguity. As KnowNo cannot predict such information, it is excluded from this evaluation. Both \ourmodel-prompt and \ourmodel-fine-tune achieve a success rate above 75\%. One can notice that the prompt baseline achieves higher performance in this specific task compared to the fine-tuned one (80\% vs 75\%).

\subsection{Real World Policy Learning}

\begin{figure}[tb]
    \centering
    \includegraphics[width=0.32\linewidth]{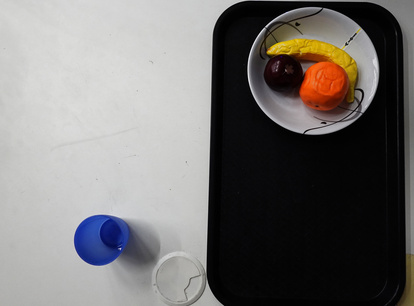}%
    \includegraphics[width=0.32\linewidth]{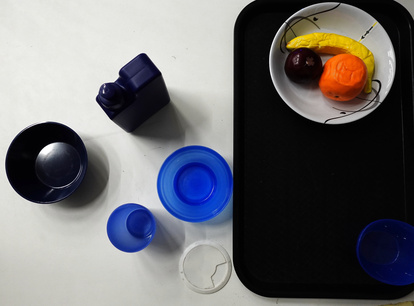}%
    \includegraphics[width=0.32\linewidth]{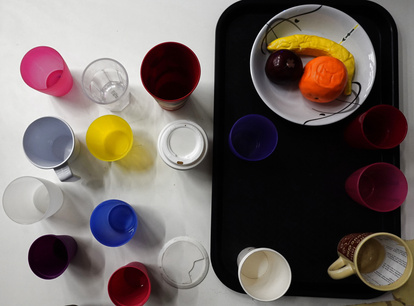}\\
    \includegraphics[width=0.32\linewidth]{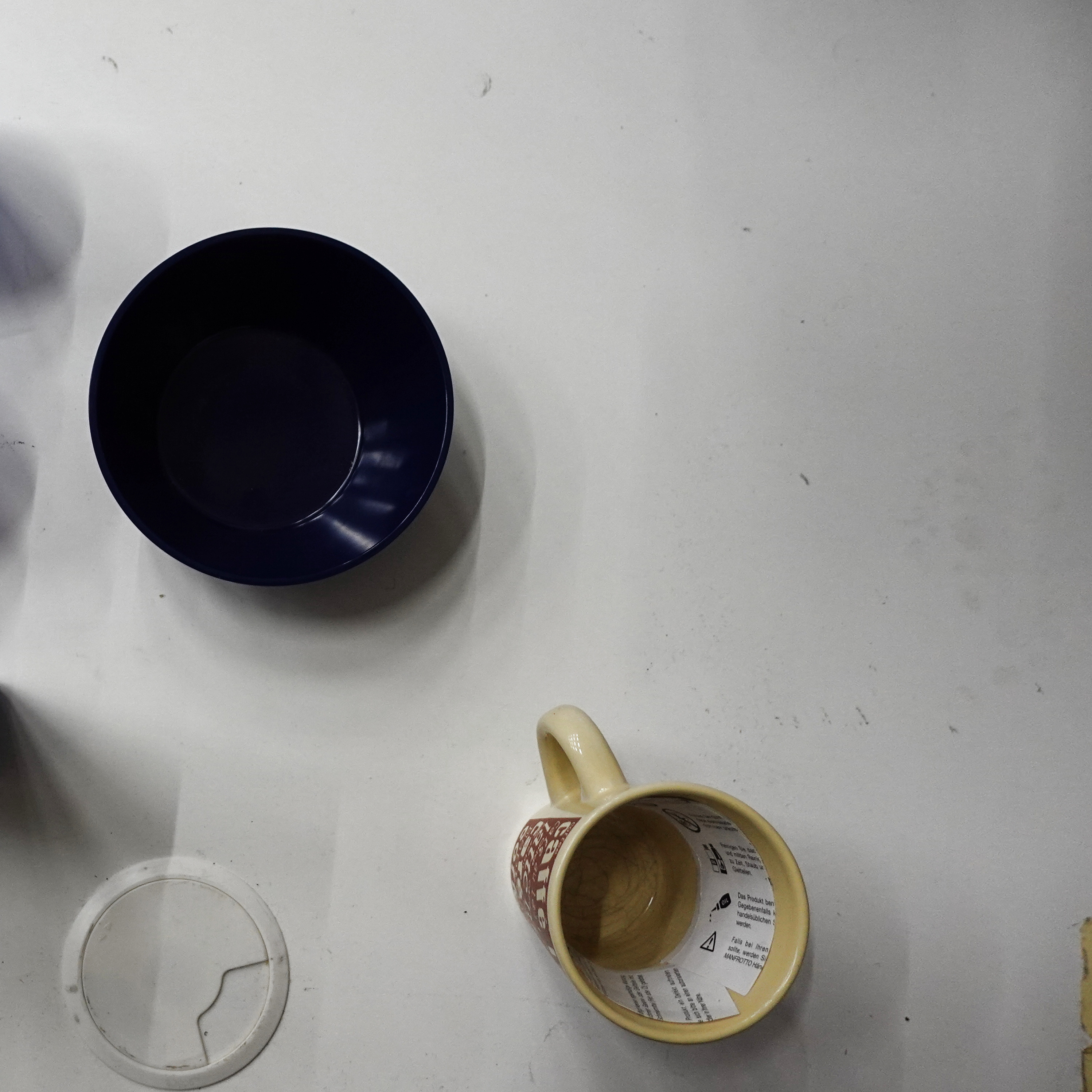}%
    \includegraphics[width=0.32\linewidth]{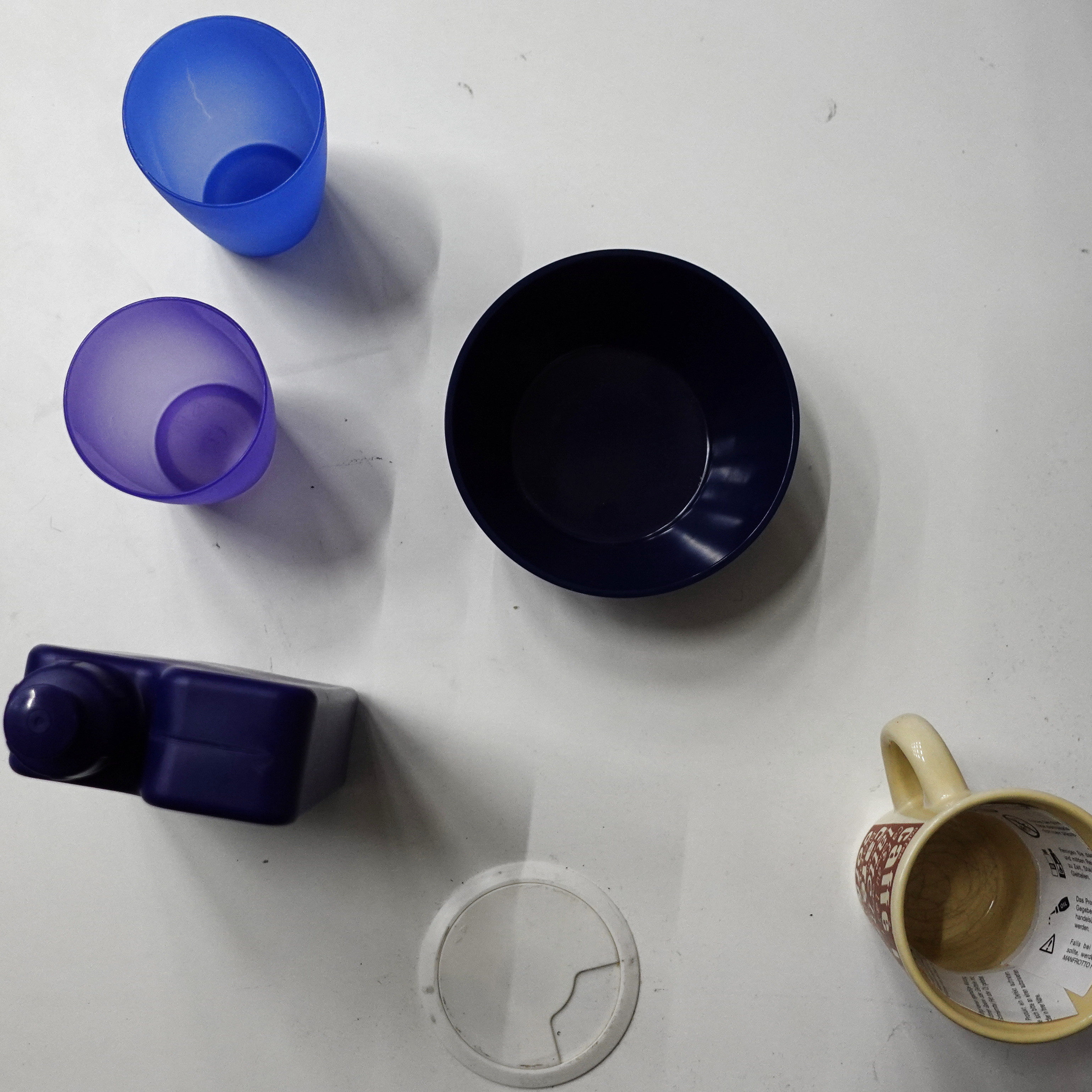}%
    \includegraphics[width=0.32\linewidth]{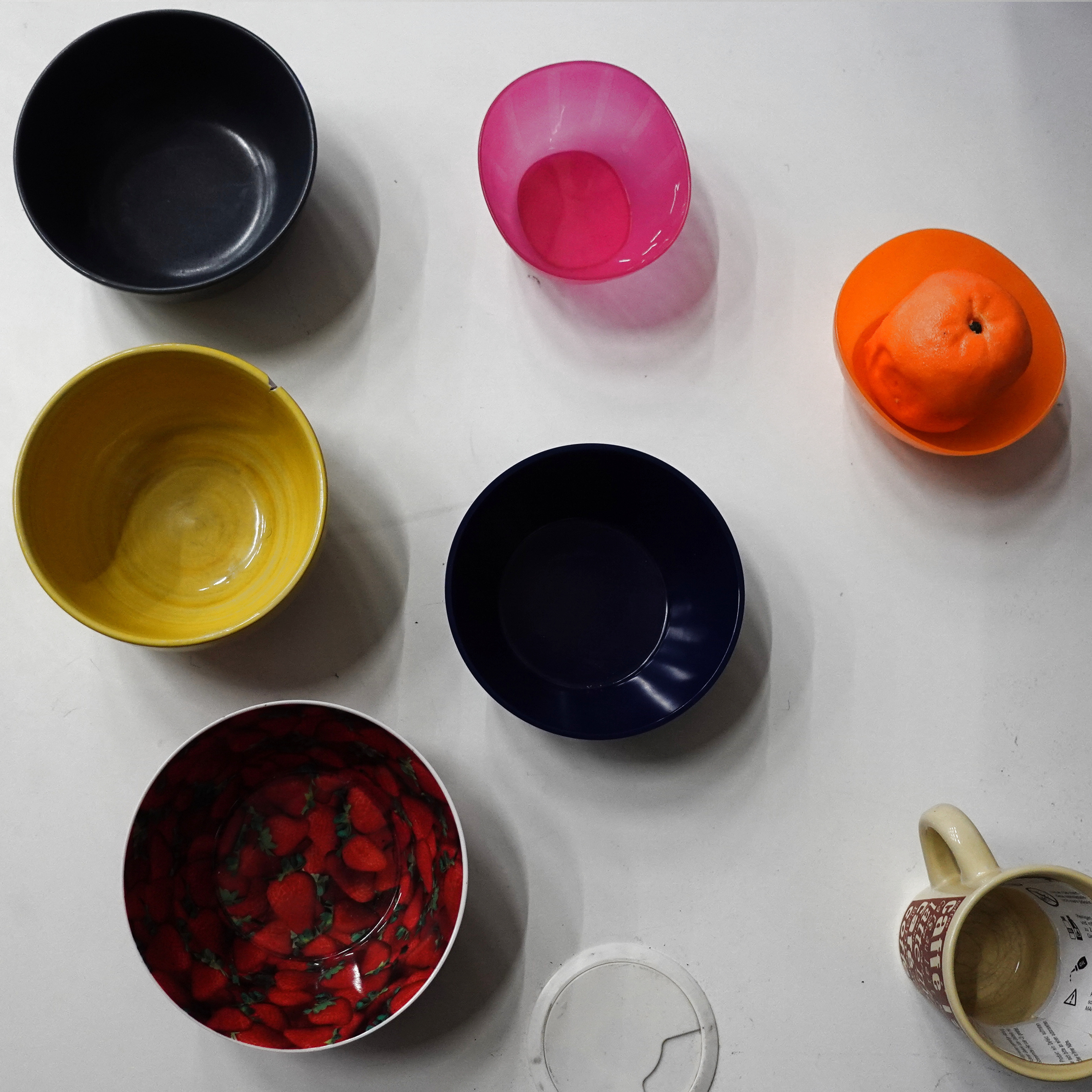}\\
    \includegraphics[width=0.32\linewidth]{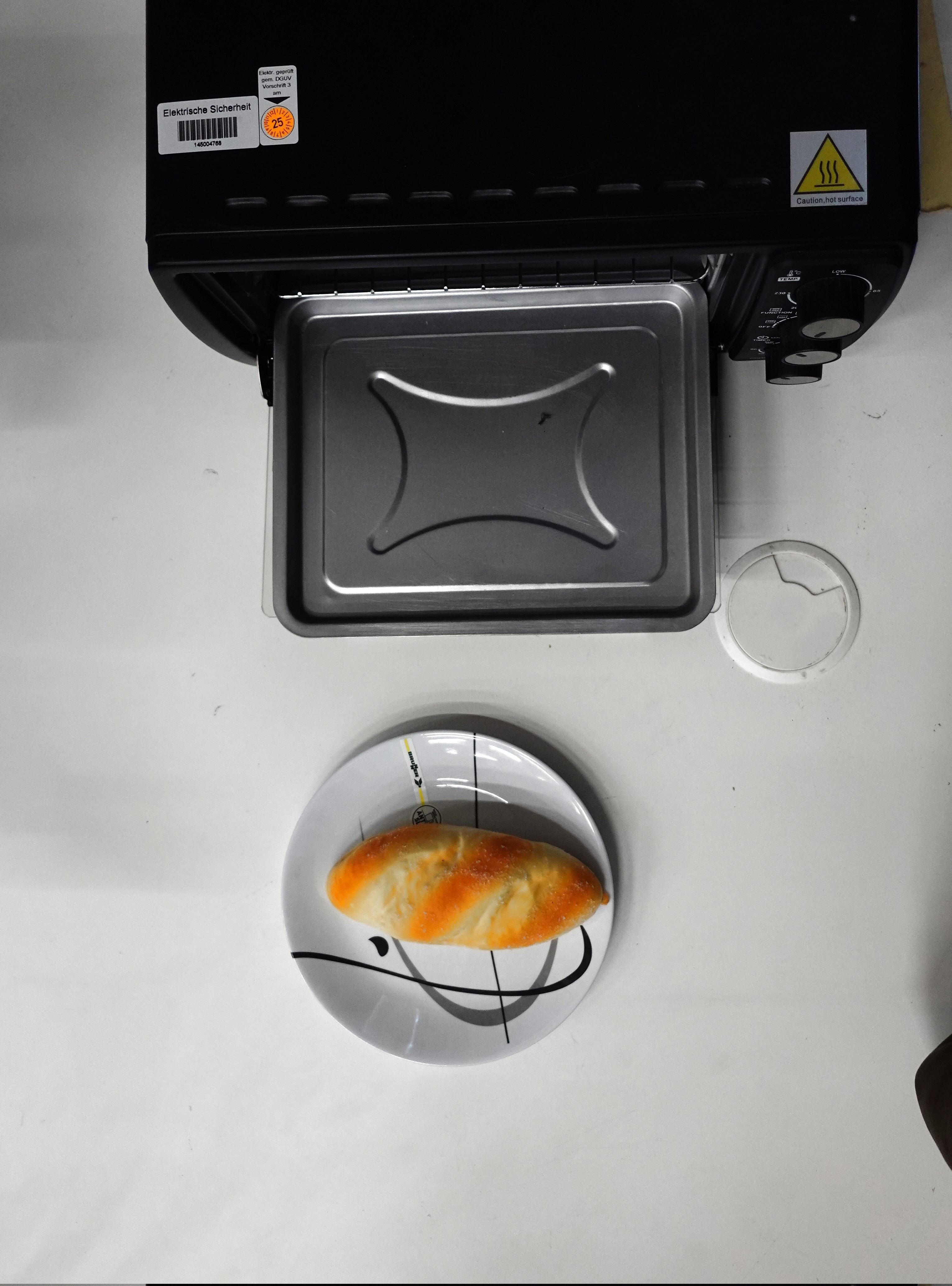}%
    \includegraphics[width=0.32\linewidth]{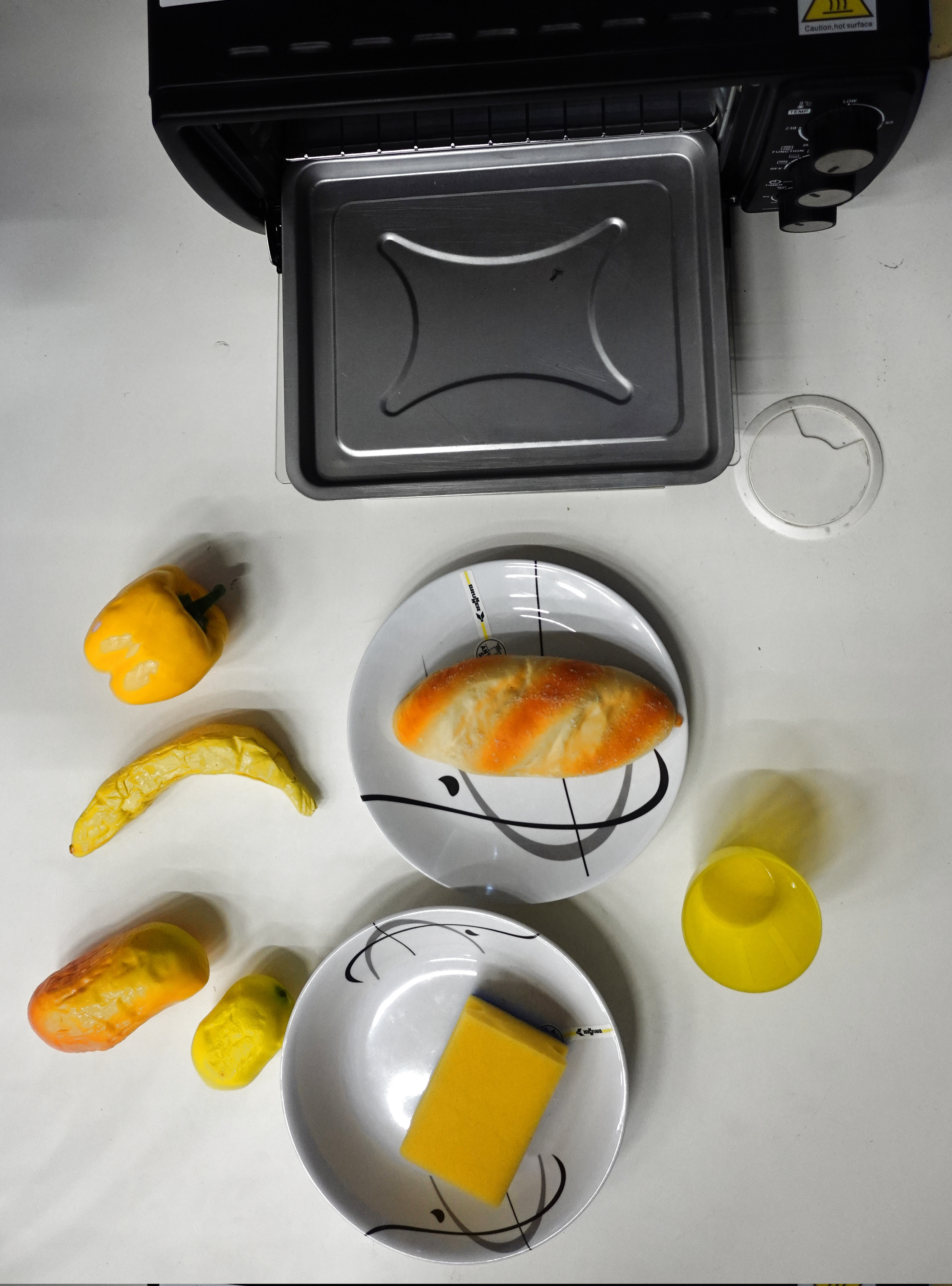}%
    \includegraphics[width=0.32\linewidth]{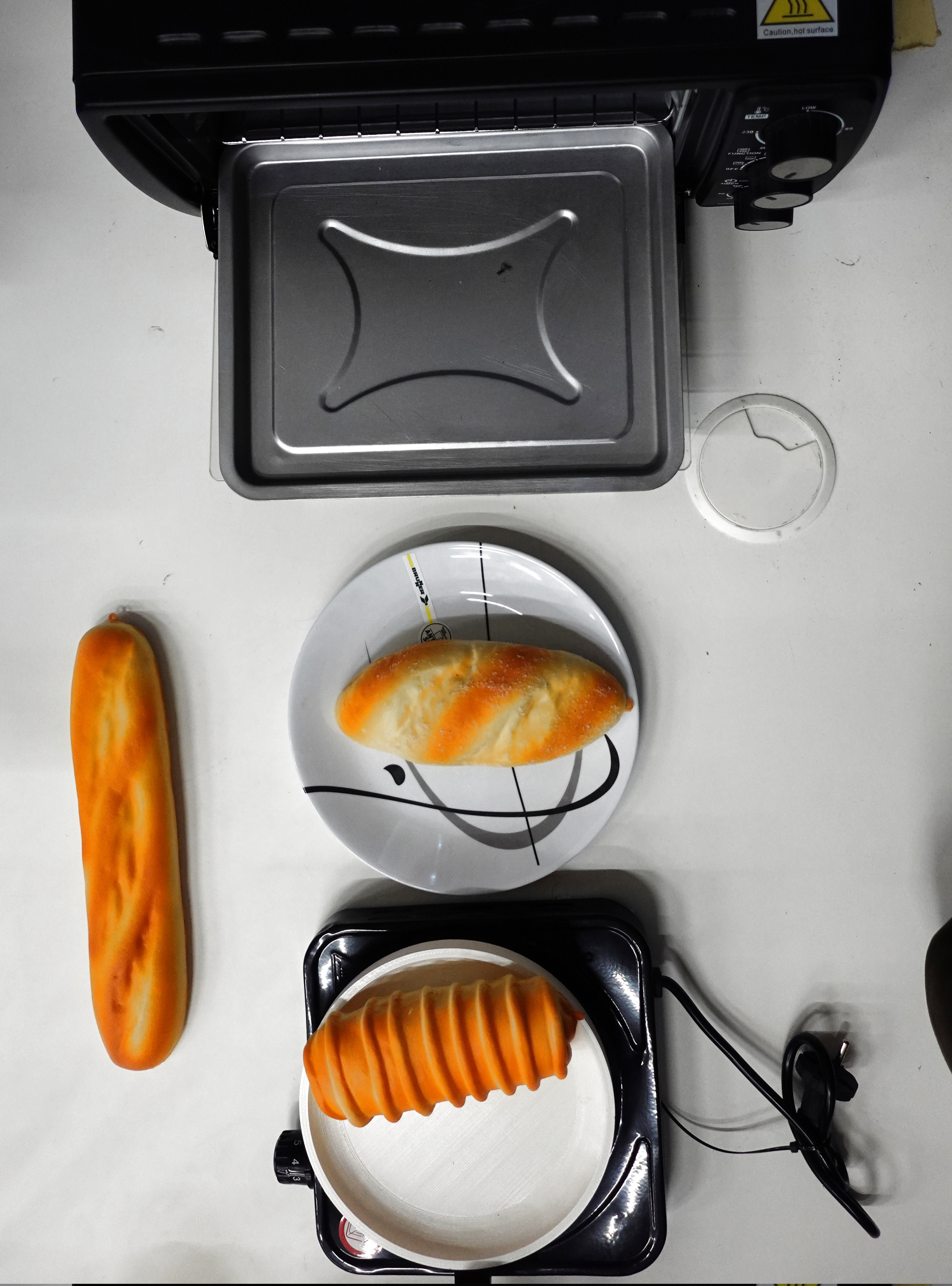}\\
    \caption{Real-world policy learning tasks.
    \textit{Top: } Place the (blue) cup on the tray.
    \textit{Middle: } Place the (blue) bowl next to the mug.
    \textit{Bottom: } Put the bread (on the plate) on the baking tray.
    Each task is evaluated under four conditions.
    \textit{Left: } with an unambiguous setup.
    \textit{Middle: } color ambiguity, with additional clutter objects of the same color.
    \textit{Right: } instance-level ambiguity, with multiple objects of the same class.
    Here, we distinguish two cases.
    First, additional object instances are present, but we still request the original objects.
    I.e., the additional instances serve only as distractors.
    Second, we also request a different object, e.g., the yellow cup instead of the blue cup.
    }\label{fig:tasks}
\end{figure}

\setlength{\tabcolsep}{4.3pt} 
\begin{table*}[t]
\centering
\caption{Object grounding success rate in the real-robot policy learning experiments.}
\vspace{-0.2cm}
\label{tab:policy-detection-success}
\begin{tabular}{l| cccc| cccc| cccc |c}
\toprule
 & \multicolumn{4}{c|}{Place the Cup on the Tray} & \multicolumn{4}{c|}{Place the Bowl next to the Mug} & \multicolumn{4}{c|}{Put the Bread on the Baking Tray} & \\
 \cmidrule(lr){2-5}\cmidrule(lr){6-9}\cmidrule(lr){10-13}
 Model & \makecell{Distinct \\ Objects} & \makecell{Color \\ Clutter} & \makecell{Instance \\ Clutter} & \makecell{Instance \\ Transfer} & \makecell{Distinct \\ Objects} & \makecell{Color \\ Clutter} & \makecell{Instance \\ Clutter} & \makecell{Instance \\ Transfer} & \makecell{Distinct \\ Objects} & \makecell{Color \\ Clutter} & \makecell{Instance \\ Clutter} & \makecell{Instance \\ Transfer} & Mean\\
 \midrule
 DINO~\cite{amir2021dino, vonhartz2024art} & \textbf{100\%} & \textbf{100\%} & 90\% & 0\% & \textbf{100\%} & \textbf{100\%} & \textbf{100\%} & 0\% & \textbf{100\%} & \textbf{100\%} & 50\% & 25\% & 72.1\%\\
 DINO+SAM~\cite{ravi2024sam2} & \textbf{100\%} & \textbf{100\%} & \textbf{100\%} & 0\% & \textbf{100\%} & \textbf{100\%} & \textbf{100\%} & 0\% & \textbf{100\%} & \textbf{100\%} & 45\% & 30\% & 72.9\%\\
 AmbRes-prompt& \textbf{100\%} & \textbf{100\%} & \textbf{100\%} & \textbf{100\%} & \textbf{100\%} & \textbf{100\%} & \textbf{100\%} & \textbf{100\%} & 0\% & 0\% & 20\% & 15\% & 76.7\%\\
 AmbRes-finetune & \textbf{100\%} & \textbf{100\%} & \textbf{100\%} & \textbf{100\%} & \textbf{100\%} & \textbf{100\%} & \textbf{100\%} & \textbf{100\%} & \textbf{100\%} & \textbf{100\%} & \textbf{90\%} & \textbf{95\%} & \textbf{98.8\%}\\
\bottomrule
\end{tabular}
\end{table*}

\begin{table*}[t]
\centering
\caption{Policy execution success rate in the real-robot policy learning experiments.}
\vspace{-0.2cm}
\label{tab:policy-success}
\begin{tabular}{l| cccc| cccc| cccc |c}
\toprule
& \multicolumn{4}{c|}{Place the Cup on the Tray} & \multicolumn{4}{c|}{Place the Bowl next to the Mug} & \multicolumn{4}{c|}{Put the Bread on the Baking Tray} & \\
 \cmidrule(lr){2-5}\cmidrule(lr){6-9}\cmidrule(lr){10-13}
 Model & \makecell{Distinct \\ Objects} & \makecell{Color \\ Clutter} & \makecell{Instance \\ Clutter} & \makecell{Instance \\ Transfer} & \makecell{Distinct \\ Objects} & \makecell{Color \\ Clutter} & \makecell{Instance \\ Clutter} & \makecell{Instance \\ Transfer} & \makecell{Distinct \\ Objects} & \makecell{Color \\ Clutter} & \makecell{Instance \\ Clutter} & \makecell{Instance \\ Transfer} & Mean\\
 \midrule
 DINO~\cite{amir2021dino, vonhartz2024art} & 90\% & 90\% & 80\% & 0\% & 90\% & 90\% & 85\% & 0\% & \textbf{100\%} & \textbf{100\%} & 50\% & 25\% & 66.7\% \\
 DINO+SAM~\cite{ravi2024sam2} & \textbf{95\%} & 90\% & \textbf{95\%} & 0\% & 95\% & 95\% & 90\% & 0\% & \textbf{100\%} & \textbf{100\%} & 45\% & 30\%  & 69.6\% \\
 AmbRes-prompt & \textbf{95\%} & \textbf{100\%} & \textbf{95\%} & 90\% & \textbf{100\%} & \textbf{100\%} & \textbf{100\%} & \textbf{100\%} & 0\% & 0\% & 20\% & 15\%  & 67.9\% \\
 AmbRes-finetune & \textbf{95\%} & 95\% & \textbf{95\%} & \textbf{95\%} & \textbf{100\%} & \textbf{100\%} & \textbf{100\%} & \textbf{100\%} & \textbf{100\%} & \textbf{100\%} & \textbf{90\%} & \textbf{95\%}  & \textbf{97.1\%} \\
\bottomrule
\end{tabular}
\end{table*}

We validate \ourmodel's utility for policy learning using TAPAS-GMM~\cite{vonhartz2024art}, a state-of-the-art few-shot policy learning approach.
TAPAS-GMM is a multi-stream method, i.e., it models the end-effector trajectory in the local coordinate frames of multiple task-relevant scene objects.
These coordinate frames are called task-parameters and are automatically extracted from camera observations~\cite{vonhartz2024art}, leveraging either a keypoint predictor~\cite{vonhartz2023treachery} or pose estimator~\cite{chisari2024centergrasp}.
When using an image keypoint model, the depth image and camera parameters are used to project the 2D keypoint to a 3D keypoint, at which the coordinate frame is attached.
The original implementation of TAPAS-GMM uses a keypoint predictor based on DINO features~\cite{amir2021dino} to achieve instance-level generalization across objects~\cite{vonhartz2024art}.
Given multiple \emph{distinct} scene objects, TAPAS-GMM then automatically selects the task-relevant object.
However, both the keypoint detection and pose estimation rely on the objects in question being unambiguous and thus fail when given multiple object candidates of the same class.
Consequently, in ambiguous scenes, ambiguity resolution is essential.\looseness=-1

To this end, we effectively integrate our proposed \ourmodel{} with TAPAS-GMM as follows.
First, for a given task, we collect five policy demonstrations.
When using \ourmodel{}, these can either be unambiguous or ambiguous, i.e., they contain multiple objects of the same class.
However, for a fair comparison with the baseline method, we collect  \emph{unambiguous} demonstrations, and only introduce the ambiguity during policy roll-out.
This enables us to investigate policy performance under different conditions of unambiguity and ambiguity.
Next, we annotate the relevant task-parameters using \ourmodel{} (or the baseline model), yielding the location of the relevant objects.
While we could utilize these object locations directly as task-parameters, we instead use them to query SAM2~\cite{ravi2024sam2} and then exploit the centers of the predicted object masks as the object locations.
Using SAM ensures that the predicted relative locations are consistent across task instances.
This is a precaution, as our model was not specifically trained for predicting consistent keypoints.
Using the predicted keypoints, we fit TAPAS-GMM on the collected demonstrations.
For inference in a new scene, we again query \ourmodel{} (or the baseline) to get the location of the desired objects and thus parameterize the policy.

We compare \ourmodel{} with the original implementation of TAPAS-GMM using DINO features.
Additionally, we implement a variant that uses the DINO keypoints to query SAM, similar to our method.
Finally, we compare the prompted and fine-tuned variants of \ourmodel.
We set up the three tasks shown in \figref{fig:tasks}.
For each task, we investigate four scenarios: \texttt{Distinct Objects} (unambiguous), \texttt{Color Clutter} (color ambiguity), \texttt{Instance Clutter} (instance-level ambiguity), and \texttt{Instance Transfer} (instance-level ambiguity with transfer to a new object instance).
Each condition is evaluated for 20 episodes.
To disentangle ambiguity resolution and location prediction, we report both the object detection success rate and the policy success rate.

The detection rates in \tabref{tab:policy-detection-success} show the DINO baseline performing strongly in the unambiguous scenario and when faced with colored clutter.
Adding clutter objects slightly affects the detection quality of the tray, but this is remedied by adding SAM.
Overall, DINO discriminates clutter instances well for objects that are simple in shape and color, such as a \emph{blue cup}.
However, it struggles with visually more complex objects, such as a \emph{loaf of bread.}
Most crucially, when multiple instances of an object class are visible, it generally prefers the reference object that was used during policy training (e.g., the blue cup) over other visible object instances (e.g., a yellow cup).
Consequently, it fails to transfer to variations of the tasks, such as grasping the yellow cup, as long as the (original) blue cup is present.
Overall, it gives the user no control over which object instance it selects.

In contrast, \ourmodel{} not only generalizes to cluttered scenes but also transfers to such task variations as described.
Besides simple generalization, such as color or size, this also enables the user to specify object instances by their relationship to other scene objects.
For example, instead of taking \enquote{the bread from the plate}, the robot can now also hand over the other bread that lies inside the pan.
We show examples of this transfer in the supplementary video.
We find that the prompted version of \ourmodel{} fails to detect the baking tray. However, the fine-tuned version succeeds.

The policy success rates in \tabref{tab:policy-success} paints a similar picture to the object detection rates.
Using SAM slightly boosts policy success rates by providing location predictions that are more consistent across task instances.
In addition, the success rates of the policy reflect the success rates of the detection.
As long as objects are distinct, visually simple, and no instance transfer is required, DINO performs on par with our method.
However, DINO becomes insufficient for more complex objects such as bread, and it fails to transfer when a different object instance is desired.
\ourmodel{} solves both problems, thus empowering strong policy results across the board.
In the supplementary video, we show qualitative results, highlighting the robustness and transferability of our approach.
\section{Conclusion}
\label{sec:conclusion}

We presented \ourmodel, a novel method for task ambiguity resolution via natural language interaction. Our model grounds task descriptions in the observed scene and reasons about task ambiguities. When the task is deemed ambiguous, \ourmodel generates the appropriate query to the human user to disambiguate it. Finally, it interprets the user clarification, grounding the information again in the observed scene and enabling successful task execution.
We evaluated the effectiveness of our model on both a simulated and a real-world domain, comparing it to the state-of-the-art baseline KnowNo~\cite{knowno2023}.
While KnowNo requires priviledged information about the objects in the scene, \ourmodel can directly reason from image observations. Despite this added challenge, our model achieves results on par with the baseline.
Finally, we demonstrate how \ourmodel improves the performance of downstream policy execution through real robot experiments. We show that supporting a language-conditioned policy with our model boosts average success rate from 69.6\% to 97.1\%.

\begin{footnotesize}
    \bibliographystyle{IEEEtran}
    \bibliography{root.bib}
\end{footnotesize}


\end{document}